\documentclass[letterpaper, 10 pt, conference]{ieeeconf}  

\IEEEoverridecommandlockouts                              

\usepackage{graphics} 
\usepackage{epsfig} 
\usepackage{amsmath} 
\usepackage{amssymb}  
\usepackage{algorithm}
\usepackage{algpseudocode}
\usepackage{adjustbox} 
\usepackage{multirow} 
\usepackage{booktabs} 
\usepackage{hyperref}

\title{\LARGE \bf
Active vision for dexterous grasping of novel objects
}

\author{Ermano Arruda        \and
        Jeremy Wyatt \and Marek Kopicki 
\thanks{We gratefully acknowledge support of FP7 grant IST-600918, PacMan, and a studentship from Brazilian Science without Borders for Ermano Arruda.}
\thanks{Arruda, Kopicki and Wyatt at
		CN-CR, 
		University of Birmingham, 
		Edgbaston, Birmingham, United Kingdom, B15 2TT, 
		Tel.: +44-121-4144788, Fax: +44-121-4144281, 
		{\tt\small  {\{exa371,msk, jlw\}@cs.bham.ac.uk}}}%
}

\begin{document}

\maketitle
\thispagestyle{empty}
\pagestyle{empty}

\begin{abstract}

How should a robot direct active vision so as to ensure reliable grasping? We answer this question for the case of dexterous grasping of unfamiliar objects. By dexterous grasping we simply mean grasping by any hand with more than two fingers, such that the robot has some choice about where to place each finger. Such grasps typically fail in one of two ways, either unmodeled objects in the scene cause collisions or object reconstruction is insufficient to ensure that the grasp points provide a stable force closure. These problems can be solved more easily if active sensing is guided by the anticipated actions. Our approach has three stages. First, we take a single view and generate candidate grasps from the resulting partial object reconstruction. Second, we drive the active vision approach to maximise surface reconstruction quality around the planned contact points. During this phase, the anticipated grasp is continually refined. Third, we direct gaze to improve the safety of the planned reach to grasp trajectory. We show, on a dexterous manipulator with a camera on the wrist, that our approach (80.4\% success rate) outperforms a randomised algorithm (64.3\% success rate).

\end{abstract}

\section{Introduction}

Grasping of novel objects is a hard problem on which there has been steady progress \cite{kopicki2015,kopicki2014a,hjelm2014a,rietzler2013a,detry2013c,stark2008a,curtis2008a,ben-amor2012a,saxena2008b,bohg2011b}. We now possess methods that are able to generate dexterous grasps for unfamiliar objects, using incomplete object reconstructions. Nonetheless, the reliability of grasping declines as the quality and completeness of the reconstruction deteriorates. Given an active vision system, we would like to minimise the number of views taken, while maximising grasping reliability. 

One approach to this would be to guide active vision to reconstruct the complete object, and for a cluttered scene the objects around it. This would be similar to the problem of active SLAM. In this paper we instead take the approach of directing gaze only to reconstruct as much of the scene as is necessary to make the executed grasp more reliable. We do this in two ways. First, we use candidate grasps to guide vision. Second, when a good grasp is found we direct vision to fill in unseen volumes of the workspace so as to make the reach to grasp more reliable. So, in each stage active vision is driven by the task. And only when a grasp is not found at a given time step we fall back on classical active SLAM.

First, we describe related work, and then proceed to describe our active vision method. This has two parts, a routine driven by the planned contact points, and a routine driven by the need to ensure a safe reach to grasp trajectory. We then present experimental results on 14 novel objects, comparing our method with a randomised view planner, and with a complete reconstruction.

\begin{figure}[t]
\centering
\includegraphics[width=0.45\textwidth]{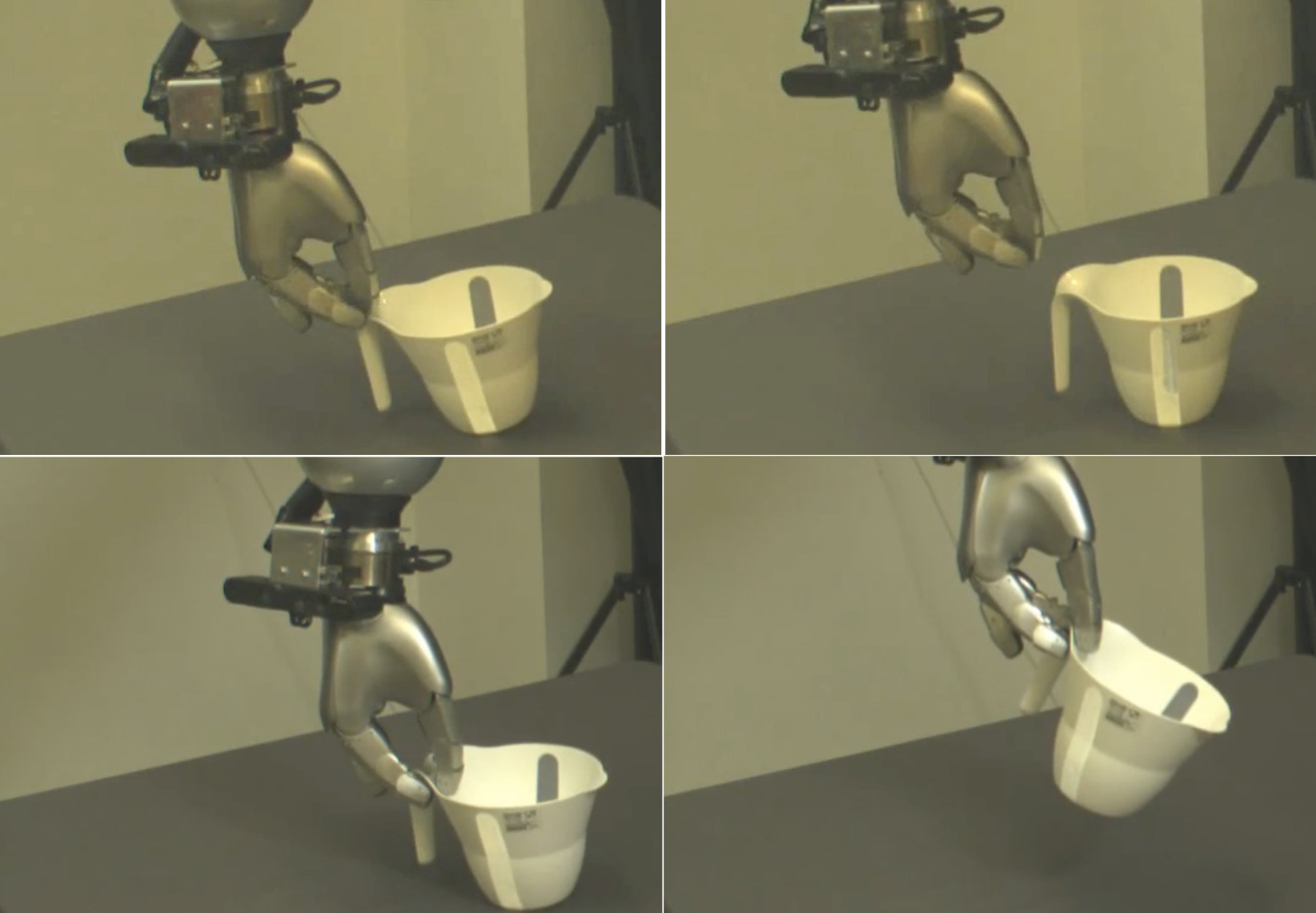}
\caption{\label{fig:fail_vs_success} Grasp failure and grasp success. The top row shows a failed grasp without active view selection. The bottom row demonstrates a successful grasp after active view selection. The difference was due to the quality of surface reconstruction close to the planned grasp points.}
\end{figure} 

\section{Related Work}

Active vision for robotics, or more generally active perception, concerns modelling, planning and control strategies for perception when the sensor can be actively moved \cite{Bajcsy88activeperception, Aloimonos1988}. The greatest advantage of active perception is that many problems that are hard to solve in the passive observer paradigm become easier. Active perception has also been applied to object recognition \cite{Atanasov_TRO14}, and to touch for grasping under uncertainty \cite{sommer2014bimanual,zito2013sequential}. For visually guided manipulation, researchers have devised strategies for view selection based on recovery of the full shape of the object to be grasped \cite{krainin2011autonomous,Chen2011IJRoboticRes}. 

Full object reconstruction is unnecessary, since most of the time only a small portion of the object surface is in contact during a grasp. In addition, recent advances have enabled grasping of novel objects in the face of incomplete reconstruction \cite{kopicki2015,kopicki2014a}. Hjelm \cite{hjelm2014a} focussed on task and grasp transferability from limited training data, i.e. demonstration and partial object point clouds. Detry \cite{detry2013c} enabled learning of grasps by letting the robot autonomously explore and try grasps while at the same time being able to transfer those self-discovered grasps to novel objects. In \cite{saxena2008b}, efforts were made towards finding stable grasps given limited visibility of object shape from cluttered scenes. The problem of shape incompleteness was tackled by Bohg et al. \cite{bohg2011b} as a problem of trying to fill the gaps between missing parts by using symmetry assumptions. 

Although there has been progress in grasping under partial reconstruction and novel objects, there exists a clear need for active perception so as to gather information most effectively. In addition, we wish to ensure robot safety, avoiding hardware damage due to unexpected collisions. We now describe our proposed approach.

\section{View Selection}

\begin{figure}[t]
\centering
\includegraphics[width=0.45\textwidth]{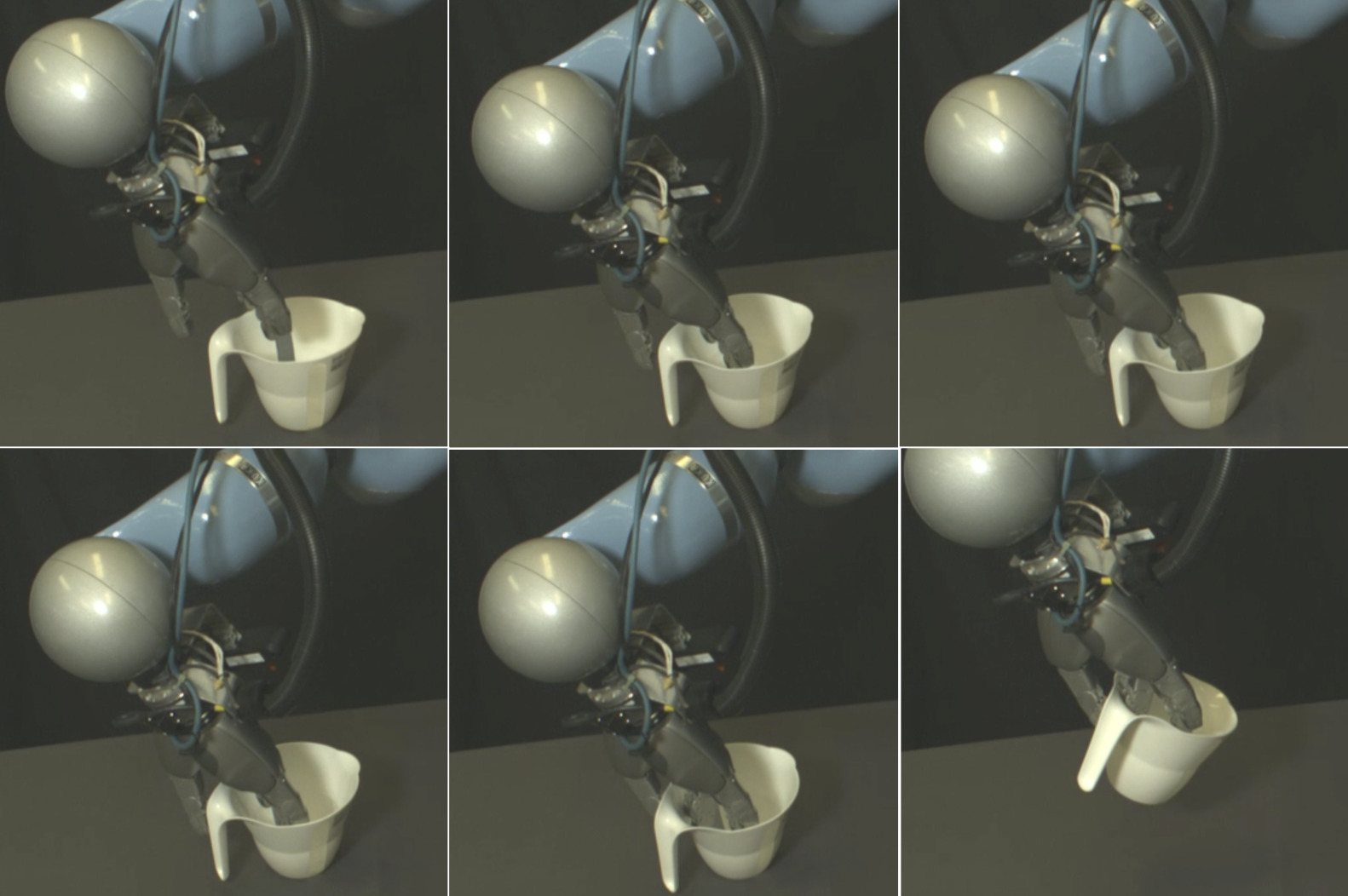}
\caption{\label{fig:jug_unsafe} An example grasp using random views. The grasp is successful. However, it can be seen that the grasp trajectory starts pushing the object aside with its fingers long before the final grasp closure takes place in the last picture on the right. This is a typical scenario that leads to failed grasps.}
\end{figure}

We first sketch our method, and then proceed to the details. The robot begins by taking a single view from a fixed location of the scene. A depth camera mounted on the robot's wrist is used. The robot is then able to choose views, which in turn provide incomplete point clouds of the object. A dexterous grasp planning algorithm is then run, which generates a large number of candidate grasps on the partial point cloud for the object. These grasps will typically assume the existence of graspable surfaces on both sides of a surface defined by the point cloud. The predicted contact locations are then used to drive the next view. The next view is chosen to maximise the quality of the point cloud at the planned contact locations. If a grasp cannot be found, we employ information gain view planning, using a 3D occupancy map. Once the quality of the relevant surface reconstruction is sufficiently high, or a limit on the number of views is reached, the grasp is fixed. Then the second phase of active vision aims to verify a safe path to the grasp location. To achieve this we again use the 3D occupancy map. This is used to calculate the probability of a collision-free trajectory. Active views for safety are driven to reduce the average entropy in cells through which the candidate reach-to-grasp trajectory passes. This ensures a safe grasp. We now proceed to describe the representations, and the three criteria used to drive active vision at different stages (contact based, information gain, and safety based).

\subsection{Representations}

\begin{algorithm}[t]
\caption{Next Best View Exploration}
\label{alg:nbv_exploration_phase1}
\begin{algorithmic}[1]
\Function{nextBestView}{$\Xi,\Gamma, \Lambda, G, V, T$}
    \State $\Omega = \emptyset$ \Comment{Most recent found contact points}
    \State $\tau = None$ \Comment{Most recent found grasp trajectory}
    \State $stop = false$
    \While {not $stop$}
    		\State $\overset{*}{\xi} = selectNBV(\Xi, \Gamma, \Lambda, V, \Omega, \tau)$
    		\State $V = append(V,\overset{*}{\xi})$ \Comment{Appending $\overset{*}{\xi}$ to V.}
        \State $\gamma = capture(\overset{*}{\xi})$ \Comment{Point cloud from pose $\overset{*}{\xi}$.}
        \State $\Gamma =\Gamma \uplus segmented(\gamma)$
        \State $\tau, \Omega = findGrasp(\Gamma, G)$ \Comment{Grasp planning with current $\Gamma_t$ based on Kopicki, Wyatt et al \cite{kopicki2015}.}
        \State $\Lambda = updateOcTree(\Lambda, \gamma, \overset{*}{\xi}, \Omega)$
        \State $T = append(T,\tau)$
        \State $stop=$ \Call{checkStop}{$V,T$}
	\EndWhile
	\State $\overset{*}{\tau} = \arg\min_{\tau \in T }  p(\tau|\Lambda)$
	\State Return $(V, \overset{*}{\tau}, \Gamma, \Lambda, T)$
\EndFunction
\Function{checkStop}{$V, T$}
	\State Return $(|V| \geq 2$ and $|T| \geq 1)$ or $|V| \geq 7)$
\EndFunction
\end{algorithmic}
\end{algorithm}

We start by describing the underlying representations used to define our approach. Let $\Xi = [\xi_1, \xi_2, \dots, \xi_N]$ be a list of possible camera poses, where $\xi_i \in SE(3)$, and $V \subset \Xi$ is the set of already visited camera poses. This list must be finite, and should provide good coverage of the workspace, but in principle views can be added to it on the fly. We describe how we picked this set in Section~\ref{sec:experiments}. In addition, let $\gamma$ be a point cloud obtained from a certain camera pose $\xi$. We define $\Gamma_t$ as the combined object point cloud, segmented from the table plane after $t$ views have been taken,
\begin{equation}
\label{eq:cloud_integration}
\Gamma_t = \Gamma_{t-1} \uplus segmented(\gamma), 
\end{equation}
i.e., $\Gamma_t$ is the result of segmenting the object point cloud from the table plane in $\gamma$ and integrating this result with our previous obtained object point $\Gamma_{t-1}$. 

In addition to the object point cloud, we also maintain a representation of the full robot workspace as a 3D occupancy grid, implemented with an octree. We shall refer to this 3D occupancy grid as $\Lambda$, which is updated after each view and observation $(\xi, \gamma)$. The implementation we use \cite{Hornung2013} allows us to easily represent known and unknown parts of the robot workspace $\Lambda$ and thus to define the information gain and safety based view planning strategies. 

It is possible to find a grasp trajectory $\tau$ by transferring a learnt grasp $G$ to the given object represented by $\Gamma_t$ using the method of Kopicki, Wyatt et al. \cite{kopicki2015}. This generates, for any partial or complete point cloud, a set of several hundred candidate grasps. These candidate grasps are generated from a statistical model learned from a single example of a particular grasp type, e.g. a pinch or power grasp. The grasps are ranked in order of likelihood according to the generative model. We pick the first ranked grasp. Then we extract the planned contact points from $\Gamma_t$, yielding a list of contacts $C = [\mathbf{c_1}, \dots, \mathbf{c_M}]$, where $\mathbf{c_i}=(w_i,\mathbf{p_i},\mathbf{n_i})$ is composed of a weight $w_i \in \mathcal{R}$, indicating its relevance to the grasp, the contact location $\mathbf{p_i} \in \mathcal{R}^3$, and the surface normal at that point $\mathbf{n_i} \in \mathcal{R}^3$. Points on the surface of the object are relevant to a grasp if they are close to a planned contact point. The weight $w_i$ or importance of each point falls off exponentially as the distance from the planned finger position increases. Let us also define $\Omega = [(\mathbf{c_1},z_1), \dots, (\mathbf{c_M},z_M)]$ as a list of contact points expanded to include the current quality $z_i$ of the observation of each point from the best view $\xi$  to date. Contact-driven vision prioritises looking at the planned contact points for which there is currently low-quality reconstruction, rather than elsewhere on the object. We now describe contact-based view selection in detail.

\begin{figure}[t]
\centering
\includegraphics[width=0.48\textwidth]{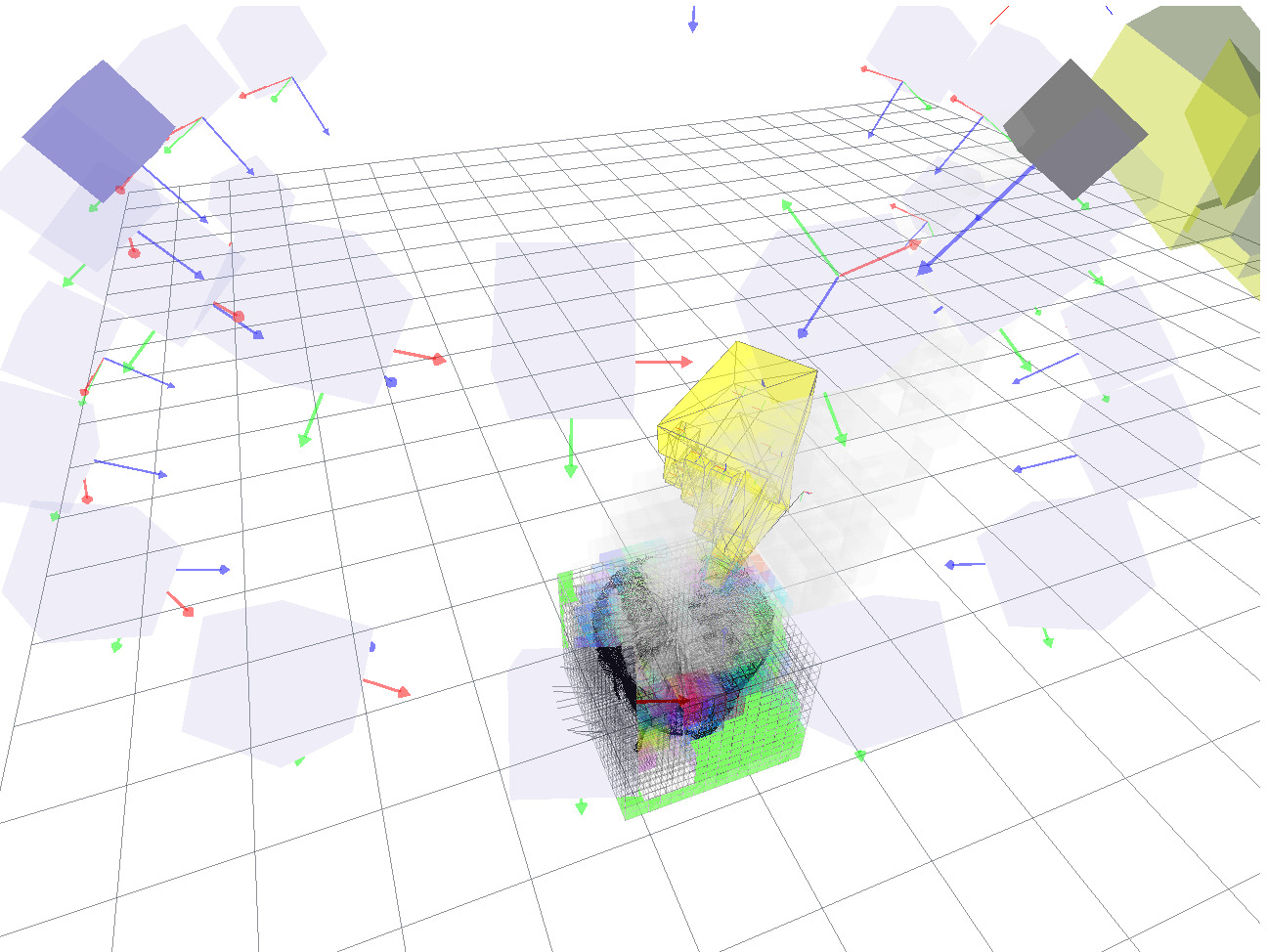}
\caption{\label{fig:view_inf_gain} View camera poses forming the set $\Xi$, camera pose highlighted in darker purple belongs to the set of visited poses $V$. The object in the centre is circumscribed by a voxelised cube. Information gain view exploration sculpts this cube when no contacts are found.}
\end{figure} 

\subsection{Contact Based View Selection} 

We let the viewing direction of a certain view pose $\xi_k \in \Xi$ be the vector $\mathbf{v_k}$, which we always constrain to point towards the object $\Gamma_t$.  We define the quality of observation of a contact point $\mathbf{c_i}$ from a given $\xi_k$ as
\begin{equation}
\label{eq:value}
\theta_{ki} = \theta(\xi_k,c_i) = \arccos(min(0,\mathbf{v_k}^T\mathbf{n_i})).
\end{equation}
This models the fact that the depth errors rise as the surface becomes perpendicular to the image plane. Thus when looking at contact point with surface normal $\mathbf{n_i}$, we assign higher values to views in which the image plane normal and the surface normal directly oppose one another, i.e. $\mathbf{n_i}$ and $\mathbf{v_k}$ form an angle of 180 degrees, according to our convention that $\mathbf{v_k}$ always looks towards the object. 

\begin{figure}[t]
\includegraphics[width=0.45\textwidth]{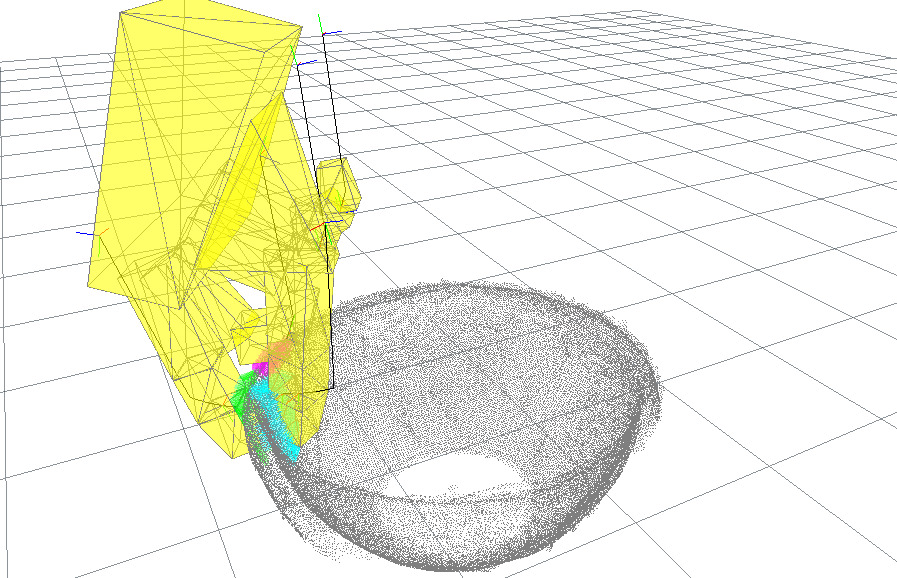}
\caption{\label{fig:contact-based} (Left) An example grasp, showing the contact regions for different finger links in different colours. Contact-driven vision attempts to view all these planned contact locations.}
\end{figure}

Thus, for each element $(\mathbf{c_i},z_i) \in \Omega$ we store the contact points $\mathbf{c_i}$, and define $z_i$ the best quality of observation to date with respect to all visited poses as 
\begin{equation}
\label{eq:z_quality}
z_i =   \arg\max_{\xi_j \in V}  \theta_{ji}.
\end{equation}

Finally, let $F_{\tau} = [\mathbf{f_1}, \dots, \mathbf{f_R}]$ be the list of finger link normals for the finger surfaces that are involved in the grasp. These are calculated for the last time step of the trajectory $\tau$. We then define the value of a particular (untried) view with respect to a particular contact $\mathbf{c_i}$ as

\begin{equation}
  \label{eq:sigma}
  \sigma(\xi_k, F_{\tau},\mathbf{c_i})= w_i \sum\limits_{r=1}^R max(\theta_{ki},z_i) \frac{1 - sign(\mathbf{f_r}^T\mathbf{n_i})}{2}.
\end{equation}

This defines high-value views as being those views which gaze head on at contact points. Note that when looking at a certain contact point $\mathbf{c_i}$ we are able either to improve our previous best viewing quality if $\theta_{ki} > z_i$, or leave it as it is. Note also that the multiplying term $\frac{(1 - sign(\mathbf{f_r}^T\mathbf{n_i})}{2}$ serves as a switch that yields $0$ or $1$. This simply ensures that to give information relevant to a particular planned contact the view must be on the side of the point cloud where the finger will contact. It does this by modelling the geometric relationship that a link must have relative to the object surface, i.e. the surface normal $\mathbf{f_r}$ of a given finger link must point in the opposite direction to the surface normal $\mathbf{n_i}$ of the object at the contact point. Finally, the normalised weight $w_i$ scales this value according to its overall relevance to the grasp as defined by the approach of Kopicki, Wyatt et al. \cite{kopicki2015}. It follows that the total utility of a given view $\xi$ is given by

\begin{equation}
\label{eq:view_contact_utility}
u_1(\xi,\Omega,\tau) = \sum\limits_{i=1}^N \sigma(\xi, F_{\tau},\mathbf{c_i}).
\end{equation}

We are then able to rank the potential views by calculating the total value of a view with respect to all contact points, and picking the view that has the maximum value according to Eq \ref{eq:view_ranking}. The algorithm will thus trade off between being able to gain information about more contacts, and gaining a great deal of information about a single contact. This proves useful when, for example, a single digit is planned to be placed on a back-surface that has not yet been observed.\footnote{This ability is due to the fact that for any thin point cloud defined surface, the grasp planning algorithm assumes that this infinitesimally thin surface can be grasped from both sides. A view of the back side will reveal if there is in fact a separately defined back surface.} In this case, viewing the back surface will be extremely informative, and will outweigh the value of observations of several finger contacts on the already observed front surface, even though the reconstruction of that front surface is imperfect.

\begin{equation}
\label{eq:view_ranking}
\overset{*}{\xi} =  \arg\max_{\xi_k \in \Xi-V} u_1(\xi,\Omega,\tau).
\end{equation}

\begin{algorithm}[t]
\caption{Select Next Best View Contact Based}
\label{alg:view_selection_contact_based}
\begin{algorithmic}[1]
\Function{NBVContactBased}{$\Xi, \Gamma, \Lambda, V, \Omega, \tau$}
	\State $\overset{*}{\xi} = None$
	\If {$|V| = 0$}
		\State $\overset{*}{\xi} = head(\Xi)$
	\ElsIf {$\Omega \neq \emptyset$}
		\State Let $\overset{*}{\xi}$ be selected according to Eq \ref{eq:view_ranking}
	\Else
		\State Let $\overset{*}{\xi}$ be selected according to Eq \ref{eq:information_gain_selection}
	\EndIf
		
	\State Return $\overset{*}{\xi}$
\EndFunction
\end{algorithmic}
\end{algorithm}

\subsection{Information Gain View Selection}

Of course, if no grasp can be found, then grasp driven view selection cannot run. In this case, the robot should look at the workspace around the recovered point cloud. To support this we define an information gain based utility function for view selection. Intuitively, this strategy makes sense, since no contacts were found with the knowledge we have about the object shape so far, represented by $\Gamma_t$. Therefore, one should ideally adopt an exploratory behaviour to seek for new parts of the object.  

For this purpose, let $bmin(\Gamma_t), bmax(\Gamma_t) \in \mathcal{R}^3$ be the respective minimum and maximum limits of the bounding box that circumscribes the object point cloud $\Gamma_t$. We are then able to extract the set of voxels $\Lambda_{object} \subset \Lambda$ inside this bounding box. If we assume the surface of this voxelised solid box $\Lambda_{object}$ is visible from all cameras, as shown in Fig \ref{fig:view_inf_gain}. Then we can define a simple strategy to minimise the entropy about the object's shape, by selecting views that are going to have maximum predicted information gain about the voxels in $\Lambda_{object}$. Intuitively, our goal is to select views such that we gradually sculpt the solid cube, in a way that we will eventually reach a constant entropy value for this cube, due to self-occluding parts of the object, from which point no views are going to bring any more information gain.

\begin{figure}[t]
\centering
\includegraphics[width=\columnwidth]{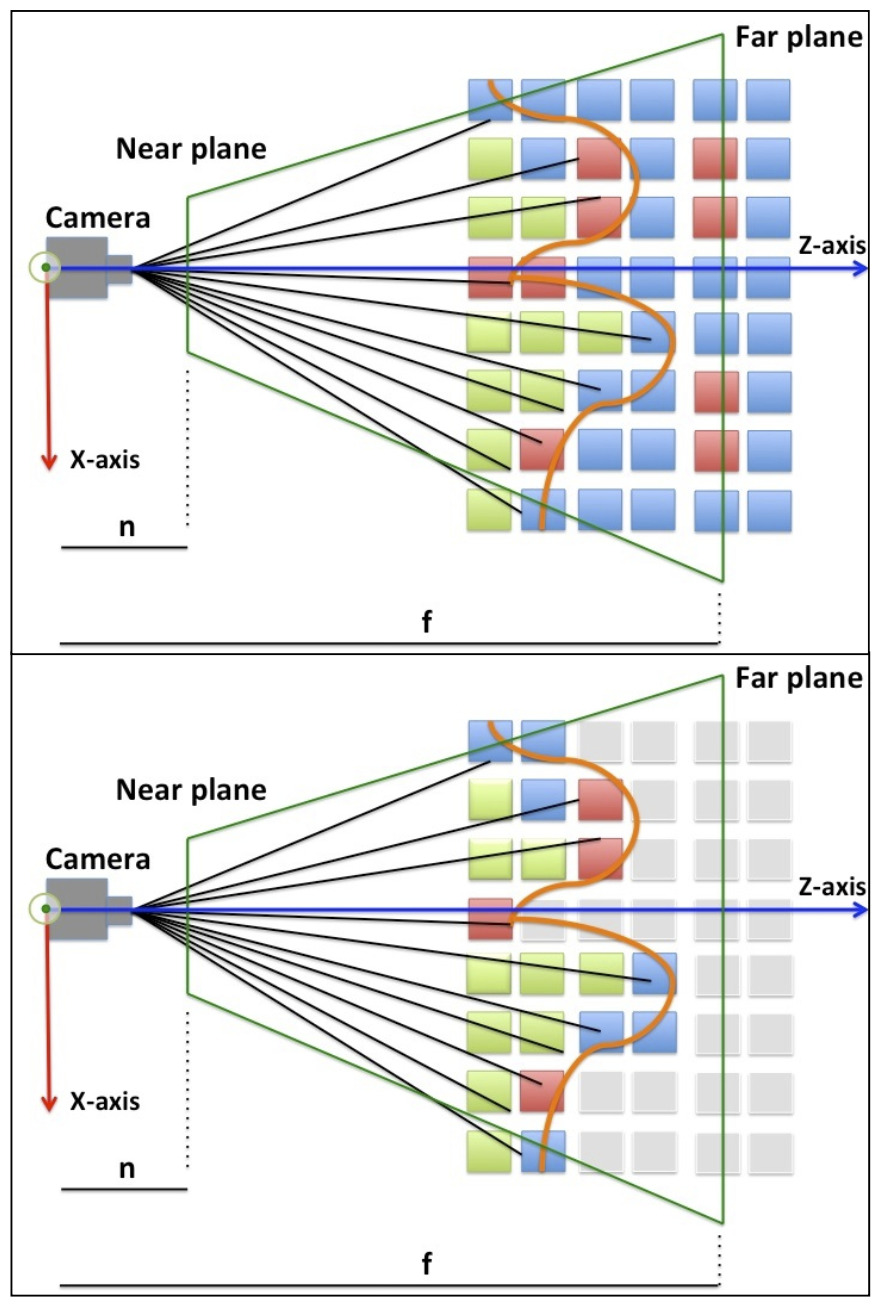}
\caption{\label{fig:voxel_visibility} Cross-section view of a typical visibility check. In the picture, occupied voxels are represented in red, free voxels are green and unknown voxels have dark-blue colour. Having defined a viewing frustum to match the real depth camera specifications, a frustum culling procedure is performed in which free voxels are assumed to be transparent, whereas unknown or occupied voxels occlude each other. As a result, only the voxels coloured on the bottom image are defined as being visible after the execution of this procedure. }
\end{figure}

Our first step to fulfilling this goal is to define a rule with which we can determine the set of visible voxels in $\Lambda_{object}$ visible from a camera pose $\xi$. The visibility test is performed using a typical frustum culling graphics procedure, with a few slight modifications. First, we transform the set of voxels $\Lambda_{object}$ into the camera coordinate system. During the projection phase of the pipeline, we allow many free voxels along the line of sight to be projected onto identical image coordinates, but we do not allow either unknown voxels, nor occupied voxels to be projected on top of one another. As a consequence, we find a border in our initial solid cube $\Lambda_{visible}(\xi) \subset \Lambda_{object}$, which contains all free voxels visible on the image plane, together with boundary voxels that might be either unknown or occupied, as shown in Fig \ref{fig:voxel_visibility}. Thus, $\Lambda_{visible}(\xi)=\{s_1, \dots, s_D\}$ is defined as our set of voxels of interest for information gain prediction. We then follow to describe the information gain prediction for the set of voxels $\Lambda_{visible}(\xi)$.

\begin{algorithm}[t]
\caption{Safety Exploration}
\label{alg:safety_exploration}
\begin{algorithmic}[1]
\Function{safetyExploration}{$\Xi, \Gamma, \Lambda, \tau, V$}
    \State $stop = false$
    \While {not $stop$}
    		\State $\overset{*}{\xi}, value = safetyNBV(\Xi, \Lambda,\tau)$
    		\State $V = append(T,\overset{*}{\xi})$
        \State $\gamma = capture(\overset{*}{\xi})$ \Comment{Point cloud from pose $\overset{*}{\xi}$.}
        \State $\Gamma = \Gamma \uplus segmented(\gamma)$
        \State $\Lambda = updateOcTree(\Lambda, \gamma, \overset{*}{\xi}, \Omega)$
        \State $T = append(T,\tau)$
        \State $stop=$ \Call{checkStopSafety}{$value$}
	\EndWhile
	\State $p = p(\tau|\Lambda)$
	\State Return $(V, p, \Gamma, \Lambda)$
\EndFunction
\Function{checkStopSafety}{$value$}
	\If {$ value \leq \beta $}
		\State Return $true$
	\Else
		\State Return $false$
	\EndIf		
\EndFunction
\end{algorithmic}
\end{algorithm}

\begin{algorithm}[t]
\caption{Safety Exploration View Selection}
\label{alg:safety_view_selction}
\begin{algorithmic}[1]
\Function{safetyNBV}{$\Xi, \Lambda, \tau$}
	\State $\Lambda_c = findPassingVoxels(\Lambda, \tau)$ \Comment{Finding voxels through which the hand is passing}
	\State Using $\Lambda_c$, let $\overset{*}{\xi}$ be selected according to Eq \ref{eq:information_gain_selection}
	\State $value = u_2(\overset{*}{\xi}, \Lambda_c)$
	\State Return $(\overset{*}{\xi}, value)$
\EndFunction
\end{algorithmic}
\end{algorithm}

\subsubsection{Information Gain Prediction}

Let the occupancy probability of a voxel $s_d \in \Lambda_{visible}$ up to our most recent observations $o_{1:t}$ be $p_{s_d}=p(s_d|o_{1:t})$. We can write the entropy of this voxel by viewing it as a Bernoulli random variable with entropy

\begin{equation}
\label{eq:voxel_entropy}
H(s_d) =  -p_{s_d}\log(p_{s_d}) - (1-p_{s_d}) \log(1-p_{s_d}),
\end{equation}

By using a log-odds representation of our occupancy probability such as in \cite{Hornung2013,Moravec85}, we can then define the future predicted occupancy probability of $s_d$ as 

\begin{equation}
\label{eq:occupancy_update}
L(s_d|o_{1:t},o'_{t+1}) = L(s_d|o_{1:t}) + L(s_d|o'_{t+1}),
\end{equation}

where $o'_{t+1} \in O=\{occupied, free\}$ is an imaginary future measurement and $L(s_d|o)$ is also called \textit{inverse sensor model} \cite{Thrun2005}. The inverse sensor model is defined likewise as in \cite{Hornung2013} as

\begin{equation}
  \label{eq:inverse_sensor_model}
  L(s_d|o) =\begin{cases}
    L_{occ}, & \text{if $o = occupied$}.\\
    L_{miss}, & \text{otherwise}.
  \end{cases}
\end{equation}

Note that our occupancy probability converted from log-odds is then 
\begin{equation}
  \label{eq:log_odds_to_prob}
  p_{sd|o_t'}=p(s_d|o_{1:t},o'_{t+1}) = 1 - \frac{1}{1 + exp(L(s_d|o_{1:t},o'_{t+1}))}.
\end{equation}

We make a simplifying assumption that an imaginary measurement has uniform distribution, i.e. $p(occupied)=p(free)=0.5$. Thus, we define our predicted entropy resulting from an imaginary measurement as the expected value 

\begin{equation}
\begin{aligned}
\label{eq:entropy_prediction}
H'(s_d) =-\sum\limits_{o' \in O} p(o')\{p_{sd|o'}\log(p_{sd|o'}) \\
				  + (1-p_{sd|o'})\log(1-p_{sd|o'})\}
\end{aligned}
\end{equation}

Therefore, the information gain of looking at a particular voxel $s_d \in \Lambda_{visible}(\xi)$ from a given view $\xi$ is given by

\begin{equation}
\label{eq:information_gain_voxel}
I(\xi,s_d) = H(s_d) - H'(s_d),
\end{equation}

where the average information gain per voxel is given by

\begin{equation}
\label{eq:information_gain_total}
u_2(\xi,\Lambda_{visible}(\xi), \Gamma_t) = \sum\limits_{s_d \in \Lambda_{visible}} \frac{I(\xi,s_d)}{D},
\end{equation}

where $D=|\Lambda_{visible}|$ is the number of visible voxels. Note that we refer to the average information gain per voxel since different views have different numbers of visible voxels in their field of view after frustum culling. This makes the predicted information different gain for different views comparable.

\subsubsection{Information Gain View Selection}

Using the definitions aforementioned, when no contacts are available, we are finally able to select next best views according to a maximum information gain strategy via

\begin{equation}
\label{eq:information_gain_selection}
\overset{*}{\xi} =  \arg\max_{\xi_k \in \Xi-V} u_2(\xi_k,\Lambda_{visible}(\xi), \Gamma_t).
\end{equation} 


\subsection{Safety View Planning}

\begin{algorithm}[t]
\caption{Grasp Driven Active Sense}
\label{alg:active_grasp}
\begin{algorithmic}[1]
\Procedure{ActiveGrasp}{$\Xi,G$}
    \State $\Gamma = \emptyset, \Lambda = \emptyset$
    \State $V = \emptyset, T = \emptyset$ \Comment{List of visited views and found trajectories, respectively. Initially empty}
    \State $grasp = false$
    \State $\overset{*}{\tau} = None$
    \While {not $grasp$}
        \State $V, \overset{*}{\tau}, \Gamma, \Lambda = nextBestView(\Xi, \Gamma, \Lambda, G, V, T)$
        \State $V, p, \Gamma, \Lambda = safetyExploration(\Xi, \Gamma, \Lambda, \overset{*}{\tau}, V)$
        
        \If {$ p \leq \alpha $}
        		\State $grasp = true$
        \Else
        		\State $T = T-\{\overset{*}{\tau}\}$ \Comment{Removing $\overset{*}{\tau}$ from our candidate trajectories}
        \EndIf
	\EndWhile
	\State $executeGrasp(\overset{*}{\tau})$
\EndProcedure
\end{algorithmic}
\end{algorithm}

In safety view planning we are interested in estimating the probability of collision prior to the execution of a given trajectory $\tau$, disregarding the collision with the final contact points $\Omega$. Effectively, we estimate the probability of an unexpected collision along the trajectory $\tau$. This is a typical scenario in which the robot hand collides with an unknown part of the object due to the fact that the collision free grasp was originally planned using an incomplete model of the object's shape $\Gamma_t$. In addition, we are also able to access how certain we are regarding this collision estimation by computing the current entropy for this particular trajectory $\tau$. As such, we select views so as to minimise the entropy of the voxels through which the robot hand is going to pass when following a given grasp trajectory $\tau$. Thus, while a path planning algorithm provides a candidate trajectory, such trajectory is not necessarily collision free, because the information acquired so far from the scene is incomplete. With a safety exploration procedure we are able to add a layer of reasoning about how much we should trust this planned path to be really free of unexpected collisions. This enables us to have a final relatively certain estimation with respect to unexpected collisions that might damage the robot hand, or simply make the grasp fail.

Let the set of voxels through which the hand bounds pass when following a trajectory $\tau$ be $\Lambda_c$. These voxels are retrieved by simulating the hand moving along the trajectory $\tau$ and querying at each time step of this trajectory the voxels the hand is passing through in our voxelised workspace $\Lambda$. Having retrieved those voxels, let $p_{s_c}$ be the probability of occupancy of a given voxel $s_c \in \Lambda_c$. The probability of collision can be calculated by

\begin{equation}
\label{eq:prob_collision}
p_{collision}(\tau, \Lambda_c) =  1 - \prod\limits_{s_c \in \Lambda_c} (1 - p_{s_c}).
\end{equation}

For numerical reasons, we prefer to refer to Eq \ref{eq:prob_collision} using only the product term, representing the probability that all voxels along the trajectory $\tau$ are free, in its logarithmic form as 

\begin{equation}
\label{eq:prob_collision_log}
\kappa(\tau, \Lambda_c) = \ln \prod\limits_{s_c \in \Lambda_c} (1 - p_{s_c}) = \sum\limits_{s_c \in \Lambda_c} \ln(1 - p_{s_c}),
\end{equation}
note that $p_{collision}(\tau, \Lambda_c) = 1 - exp(\kappa(\tau, \Lambda_c))$.

Finally, to select views in order to get better estimations for Eq \ref{eq:prob_collision}, we use the same utility function defined in \ref{eq:information_gain_total}. Thus if we let $\Lambda_{c_{visible}}(\xi) \subset \Lambda_c$ be the set of visible voxels by a certain view pose $\xi$. Next best views are then selected according to

\begin{equation}
\label{eq:information_gain_selection_collision}
\overset{*}{\xi} =  \arg\max_{\xi_k \in \Xi-V} u_2(\xi_k,\Lambda_{c_{visible}}(\xi), \Gamma_t).
\end{equation}

In practice, we allow safety exploration to run while the information gain is above a predefined threshold, i.e. $ u_2(\xi_k,\Lambda_{c_{visible}}(\xi), \Gamma_t) > \beta$. If this criteria is not met, the final probability of collision is reported according to Eq \ref{eq:prob_collision}. The trajectory $\tau$ is therefore executed or not based on the probability of collision.

\begin{figure}[t]
\centering
\includegraphics[width=0.48\textwidth]{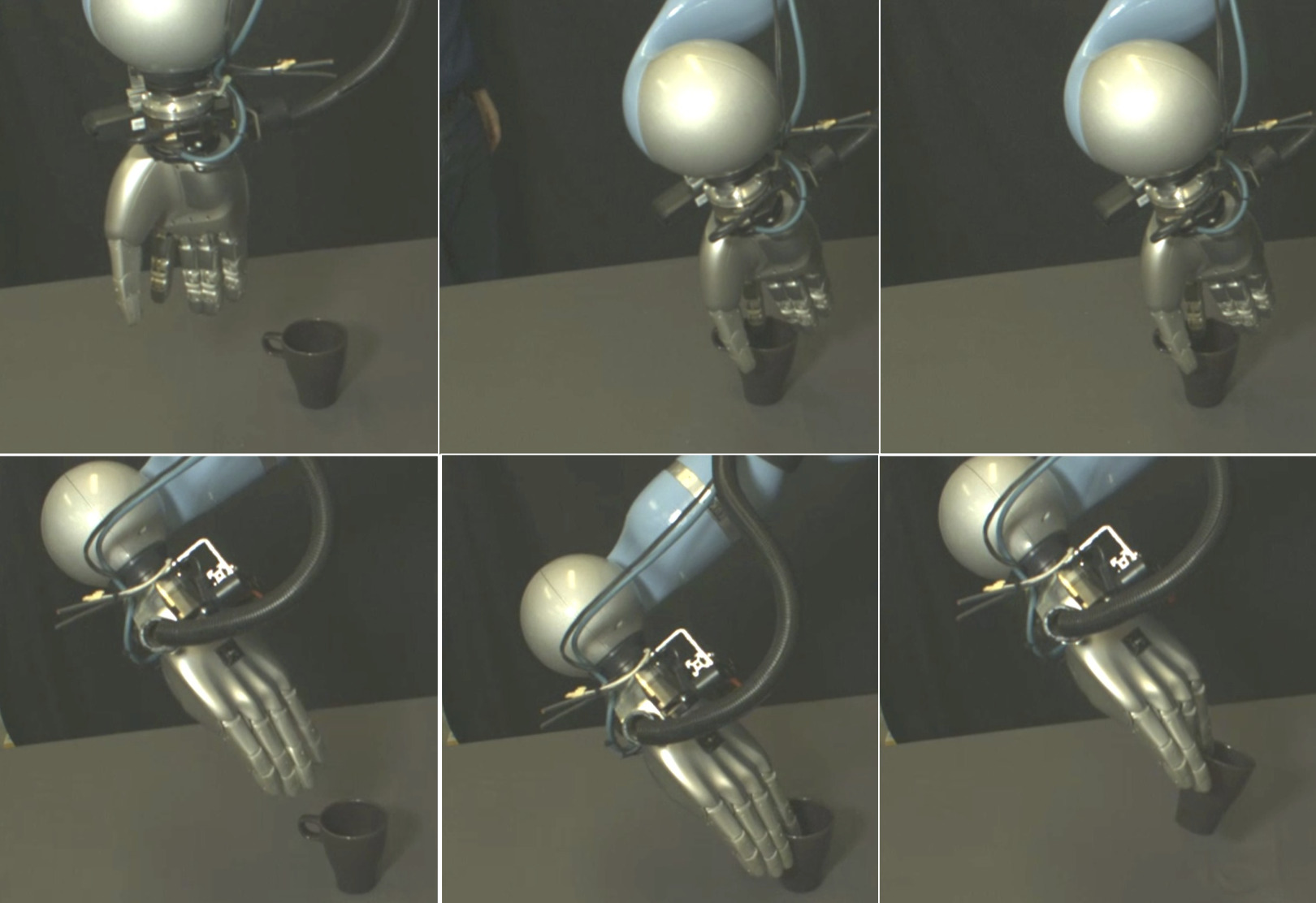}
\caption{\label{fig:random_fail_vs_nbv_success} Top three pictures show a failed grasp due to unexpected collision with parts of the object that are not involved in the grasp. Bottom three pictures show a successful and safe grasp selected by our approach. }
\end{figure} 

\section{Experiments}
\label{sec:experiments}
Pseudocode for our approach is given in  Alg \ref{alg:active_grasp}, which is divided into two sub-phases. First, a contact-based next best view exploration procedure is run as outlined by Alg \ref{alg:nbv_exploration_phase1}. In this first phase, at least two views are selected, up to a maximum of 7 views if after the second view no grasp trajectory and contacts are found. The first view is fixed, only subsequent views after this fixed view are selected according to the criteria for contact-based view selection. The second phase of Alg \ref{alg:active_grasp} is run in order to estimate the probability of collision of the most promising candidate trajectory $\overset{*}{\tau}$, selected as the trajectory with the lowest probability of collision prior to the safety view exploration phase, given our current knowledge of the object $\Gamma_t$ and workspace $\Lambda$. This second phase is outlined in Alg \ref{alg:safety_exploration}. Note that the safety exploration phase stops if the currently selected view predicts information gain below a certain threshold $\beta$.  If, at the end of the safety exploration phase, we discover that this trajectory $\overset{*}{\tau}$ has collision probability above a certain threshold $\alpha$, we reject $\overset{*}{\tau}$ and cycle back to phase 1, i.e. Alg \ref{alg:nbv_exploration_phase1}. 

\subsection{Methodology}

\begin{figure}[t]
\centering
\includegraphics[width=0.48\textwidth]{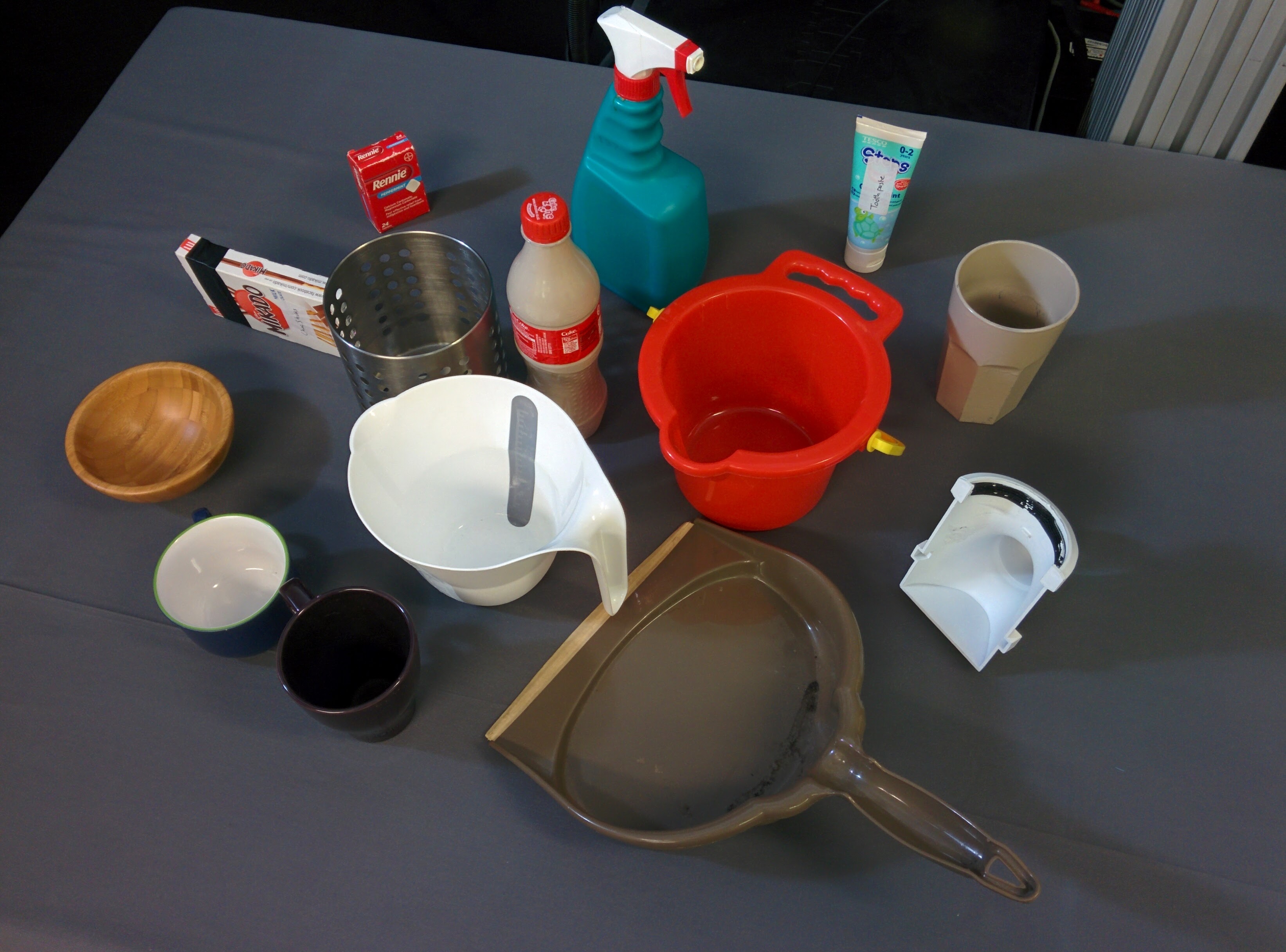}
\caption{\label{fig:trial_objects} The 14 objects used for trials.}
\end{figure}

Using Alg \ref{alg:active_grasp} we performed trials on a set of 14 novel objects shown in Fig \ref{fig:trial_objects}. A total of 4 trials per object were performed. The set of views were thirty-four roughly evenly spaced locations on two concentric view hemispheres pointing at the centre of the workspace.\footnote{Note that the algorithm can easily consider views generated on the fly. We restrict the view set here because the full visual SLAM system then required is beyond the scope of the paper.} In our experiments, we compared our algorithm with a random view selection strategy. In other words, we substituted all calls of the selection procedures Alg \ref{alg:view_selection_contact_based} and Alg \ref{alg:safety_view_selction} by a uniform random view selection scheme. Furthermore, we limited the two phases of this modified randomised approach to be constrained to the same number of views that our algorithm performed in both phases. It is also important to note that in our experiments we have set the size of the voxels in our 3D occupancy map $\Lambda$ to be $0.0025 m$, for relatively fine precision. Table \ref{tab:results} shows the final data for this experiment. In addition, we tested performance on a cluttered scene using an early version of the algorithm that employed only the grasp contact refinement strategy.

All trials were performed with the Boris manipulation platform, which was mounted with a PrimeSense Carmine 1.09 depth camera on its wrist. The robot hand utilised was the DLR-HIT2 hand \cite{HLiu2008}. A grasp was considered successful if the robot could lift the object without letting it fall or slip. 

\subsection{Results}

The results shown in Table \ref{tab:results} show the contrast between the two approaches. We first note that the success rate of our proposed view selection approach was 80.4\%, whereas the modified randomised approach showed a success rate of only 64.3\%. A closer look at Table \ref{tab:results} reveals that random exploration tended to yield unsafe grasps, under the same view number constraints as our active view selection approach. This indicates that random view selection would probably need to cycle back to generate new grasp trajectory candidates more times, which seems a natural consequence of its sub-optimal exploratory behaviour. One such example is highlighted by Fig~\ref{fig:random_fail_vs_nbv_success}, in which the final trajectory executed with a probability of collision 1.0 and, indeed, makes the robot hand collide with a part of the mug not involved in the grasp, finally failing for safety reasons. We also note that our collision probability appears to be over-sensitive, the random approach also succeeded for various cases in which the probability of collision was 1.0. Nonetheless, even for successful grasps as the one depicted in  Fig \ref{fig:jug_unsafe}, grasps with probability 1.0 tended to collide prematurely with different parts of the object. In addition, we also noted that for the case of the toothpaste, the trivial solution of a grasp with as few collisions as possible might yield grasps with very poor grip. This indicates future work towards a middle ground between these two extremes.

\begin{figure}[t]
\centering
\includegraphics[width=0.95\columnwidth]{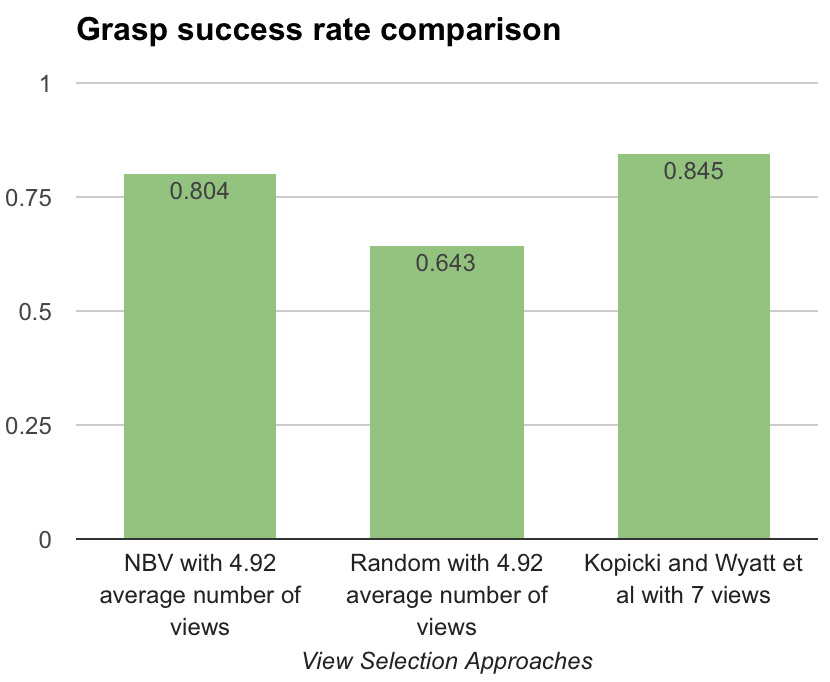}
\caption{\label{fig:success_rate_comparison} Grasp success rate comparison. }
\end{figure}

As shown by Table~\ref{tab:results} and in Fig~\ref{fig:success_rate_comparison}, our approach had competitive success rate to prior work done by Kopicki and Wyatt et al \cite{kopicki2015}. In our experiments, our approach used on average 4.92 views for grasp planning, as compared with 7 in \cite{kopicki2015}. Additional views were only used to assess safety \footnote{A video illustrating our approach can be found at \url{https://youtu.be/uBSOO6tMzwA}}. 

Finally, we tested the robot on a cluttered scene with three novel objects. This single trial completed successfully \footnote{See \url{https://youtu.be/hnGgsWtKzPw}.}.

\section{Conclusions}

We have proposed an effective approach  for view selection comprising two stages. The first stage guides gaze by planned contacts, seeking a low noise point cloud near those contacts. If no contact points are available, we guide gaze so as to minimise the entropy of the 3D occupancy grid map around our object cloud $\Gamma_t$. After candidate grasps are found, we gaze to assess the safety of the reach-to-grasp trajectory candidate. We showed that this yields a better success rate compared to a random strategy and competitive success rate to \cite{kopicki2015}, while using fewer views for grasp planning. In future work we plan to utilise visual SLAM techniques, to learn the value of information, and to exploit recent insights from active SLAM in non-manipulation tasks, e.g. \cite{Charrow-icra-15}.

\begin{table*}[!bt]
\centering
\caption{Trial Results}
\label{tab:results}
\begin{adjustbox}{width=0.95\textwidth}
\begin{tabular}{@{}llllllll@{}}
\toprule
\multicolumn{1}{|l|}{}                         & \multicolumn{1}{c|}{\begin{tabular}[c]{@{}c@{}}Phase 1 \\ Average View \\ Count\end{tabular}}          & \multicolumn{1}{c|}{\begin{tabular}[c]{@{}c@{}}Phase 2 \\ Average View \\ Count\end{tabular}}          & \multicolumn{2}{c|}{\begin{tabular}[c]{@{}c@{}} Average Success Rate \\ per Object \end{tabular}}     & \multicolumn{2}{l|}{\begin{tabular}[c]{@{}l@{}} Average Collision \\ Probability per Object \end{tabular}}                & \multicolumn{1}{l|}{\begin{tabular}[c]{@{}l@{}} Average \\ Grasp \\ Views\end{tabular}}                     \\ \midrule
\multicolumn{1}{|l|}{\multirow{2}{*}{Objects}} & \multicolumn{1}{l|}{\multirow{2}{*}{\begin{tabular}[c]{@{}l@{}}NBV \& \\ Random\end{tabular}}} & \multicolumn{1}{l|}{\multirow{2}{*}{\begin{tabular}[c]{@{}l@{}}NBV \& \\ Random\end{tabular}}} & \multicolumn{1}{l|}{\multirow{2}{*}{NBV}} & \multicolumn{1}{l|}{\multirow{2}{*}{Random}} & \multicolumn{1}{l|}{\multirow{2}{*}{NBV}} & \multicolumn{1}{l|}{\multirow{2}{*}{Random}}             & \multicolumn{1}{l|}{\multirow{2}{*}{\begin{tabular}[c]{@{}l@{}}t of\\ $\Gamma_t$\end{tabular}}} \\
\multicolumn{1}{|l|}{}                         & \multicolumn{1}{l|}{}                                                                          & \multicolumn{1}{l|}{}                                                                          & \multicolumn{1}{l|}{}                     & \multicolumn{1}{l|}{}                        & \multicolumn{1}{l|}{}                     & \multicolumn{1}{l|}{}                                    & \multicolumn{1}{l|}{}                                                                           \\ \midrule
bowl small                                     & 2.67                                                                                           & 3.67                                                                                           & 1                                         & 1                                            & 0.0006166666667                           & 0.6667388333                                             & 3.17                                                                                               \\
bucket                                         & 4.17                                                                                           & 3.34                                                                                           & 1                                         & 1                                            & 0.0000525                                 & 0.5037015                                                & 6.34                                                                                               \\
chopsticks                                     & 3                                                                                              & 3.67                                                                                           & 1                                         & 0.75                                         & 0.005303                                  & 0.0002226666667                                          & 5.67                                                                                               \\
coke                                           & 2.5                                                                                            & 4.5                                                                                            & 0.75                                      & 0.25                                         & 0.01711816667                             & 0.6674898333                                             & 3.67                                                                                               \\
cup1                                           & 3.17                                                                                           & 4.34                                                                                           & 0.75                                      & 0.5                                          & 0.001250833333                            & 0.8333813333                                             & 6.17                                                                                               \\
dustpan                                        & 3                                                                                              & 3                                                                                              & 1                                         & 1                                            & 0.005206333333                            & 1.0                                                      & 4                                                                                               \\
glass2                                         & 3.5                                                                                            & 3.17                                                                                           & 0.25                                      & 0                                            & 0.6666955                                 & 1.0                                                      & 4                                                                                               \\
guttering                                      & 3                                                                                              & 3.67                                                                                           & 1                                         & 0.5                                          & 0.005433166667                            & 0.8333818333                                             & 4                                                                                               \\
jug                                            & 3.34                                                                                           & 4.34                                                                                           & 1                                         & 1                                            & 0.07424566667                             & 0.833678                                                 & 6.34                                                                                              \\
mrmuscle                                       & 2.67                                                                                           & 3.17                                                                                           & 1                                         & 0.75                                         & 0.002464                                  & 1.0                                                      & 4.5                                                                                               \\
mug1                                           & 2.84                                                                                           & 3.67                                                                                           & 0.25                                      & 0.25                                         & 0.003175666667                            & 0.6672291667                                             & 5.34                                                                                               \\
rennie                                         & 3                                                                                              & 3.17                                                                                           & 1                                         & 1                                            & 0.004263666667                            & 0.5001438333                                             & 4.17                                                                                               \\
stand2                                         & 4.67                                                                                           & 2.34                                                                                           & 0.5                                       & 0.5                                          & 0.003219666667                            & 0.666765                                                 & 4.67                                                                                               \\
toothpaste                                     & 3.84                                                                                           & 4                                                                                              & 0.75                                      & 0.5                                          & 0.003282333333                            & 0.3352755                                                & 6.84                                                                                               \\ \midrule
                                               &                                                                                                & \begin{tabular}[c]{@{}l@{}}Overall Success \\ Rate\end{tabular}                                        & 0.804                                     & 0.643                                        &                                           & \begin{tabular}[c]{@{}l@{}}Overall \\ Grasp Views\end{tabular} & 4.92                                                                                             \\ \bottomrule
\end{tabular}
\end{adjustbox}
\end{table*}









\bibliography{../IEEEtranBST/IEEEexample,grasping}
\bibliographystyle{plain}      

\end{document}